\documentclass[letterpaper, 10 pt, conference]{ieeeconf}  %
\usepackage[left=1.91cm,right=1.91cm,top=1.91cm,bottom=1.91cm]{geometry}

\IEEEoverridecommandlockouts                              %

\usepackage{graphics} %
\usepackage{epsfig} %
\usepackage{mathptmx} %
\usepackage{times} %
\usepackage{amsmath} %
\usepackage{amssymb}  %
\usepackage{cleveref}

\usepackage[dvipsnames]{xcolor}

\usepackage{xspace}
\usepackage{comment}
\usepackage{multicol}
\usepackage{multirow}
\usepackage{graphicx}
\usepackage{algorithm}
\usepackage{algpseudocode}
\usepackage{scalefnt}
\usepackage{booktabs}

\usepackage[small,bf]{caption}
\usepackage{cite}

\title{SAID-NeRF: Segmentation-AIDed NeRF\\for Depth Completion of Transparent Objects}

\author{
Avinash Ummadisingu$^{1}$,
Jongkeum Choi$^{2}$,
Koki Yamane$^{3}$,\\
Shimpei Masuda$^{1}$,
Naoki Fukaya$^{1}$,
Kuniyuki Takahashi$^{1}$
\thanks{$^{1}$A. Ummadisingu, S. Masuda, N. Fukaya, and K. Takahashi are with Preferred Networks, Inc.
        {\tt\footnotesize 
        \{{ummavi, masuda, fukaya, takahashi\}@preferred.jp}}
        $^{2}$J. Choi is with the University of Tokyo. This work is done in a part-time job at Preferred Networks, Inc.
        {\tt\footnotesize 
        choijongkeum@g.ecc.u-tokyo.ac.jp}
        $^{3}$K. Yamane is with the University of Tsukuba. This work is done in a part-time job at Preferred Networks, Inc.
        {\tt\footnotesize 
        yamane.koki.td@alumni.tsukuba.ac.jp}
}}
\begin{document}
\maketitle
\thispagestyle{empty}
\pagestyle{empty}

\begin{abstract}
Acquiring accurate depth information of transparent objects using off-the-shelf RGB-D cameras is a well-known challenge in Computer Vision and Robotics.
Depth estimation/completion methods are typically employed and trained on datasets with quality depth labels acquired from either simulation, additional sensors or specialized data collection setups and known 3d models.
However, acquiring reliable depth information for datasets at scale is not straightforward, limiting training scalability and generalization.
Neural Radiance Fields (NeRFs) are learning-free approaches and have demonstrated wide success in novel view synthesis and shape recovery.
However, heuristics and controlled environments (lights, backgrounds, etc) are often required to accurately capture specular surfaces. 
In this paper, we propose using Visual Foundation Models (VFMs) for segmentation in a zero-shot, label-free way to guide the NeRF reconstruction process for these objects via the simultaneous reconstruction of semantic fields and extensions to increase robustness. 
Our proposed method \emph{S}egmentation-\emph{AID}ed \emph{NeRF} (SAID-NeRF) shows significant performance on depth completion datasets for transparent objects and robotic grasping.

\end{abstract}

\section{Introduction}
\label{sec:intro}

Acquiring accurate 3d information of transparent objects is a crucial but difficult challenge in Robotics and Computer Vision. 
These objects demonstrate several non-Lambertian properties leading to highly view-dependent appearance, if at all visible due to transparency, and specular effects like reflection, refraction, scattering, and dispersion complicating both visual perception and depth capture via RGB-D cameras. Alternative sensor modalities have been studied but have limitations due to cost, calibration and sensitivity ~\cite{jiang2023robotic}.

Machine learning approaches can be utilized to estimate or complete noisy or missing depth values~\cite{fang2022transcg,wang2023mvtrans,dai2022domain,sajjan2020clear}.
Deep Learning methods typically require large amounts of data, but the acquisition of accurate labels for depth is challenging due to the difficulties in accurately simulating them or complex labeling setups needed~\cite{sajjan2020clear,chen2022clearpose, fang2022transcg}.
Furthermore, enhancing the out-of-distribution generalization capabilities of these methods remains a considerable hurdle.

A contrasting approach is the use of NeRF-based methods for 3d scene reconstruction from multiple views.
NeRFs allow for a learning-free, per-scene optimization approach, sidestepping issues related to training and generalization~\cite{mildenhall2021nerf, mueller2022instant, wang2021neus}.
However, they are often limited by the quality and quantity of views supplied.
This is particularly important for the sufficient capture of transparent object surfaces from different views and involves potentially inducing specular effects like reflections, by adding additional light sources~\cite{ichnowski2021dex}. While NeRFs demonstrate impressive visual reconstructions of transparent objects, the recovered depth is often too inaccurate/noisy to use in applications like robotic grasping.

\begin{figure*}
    \centering
    \includegraphics[width=2\columnwidth]{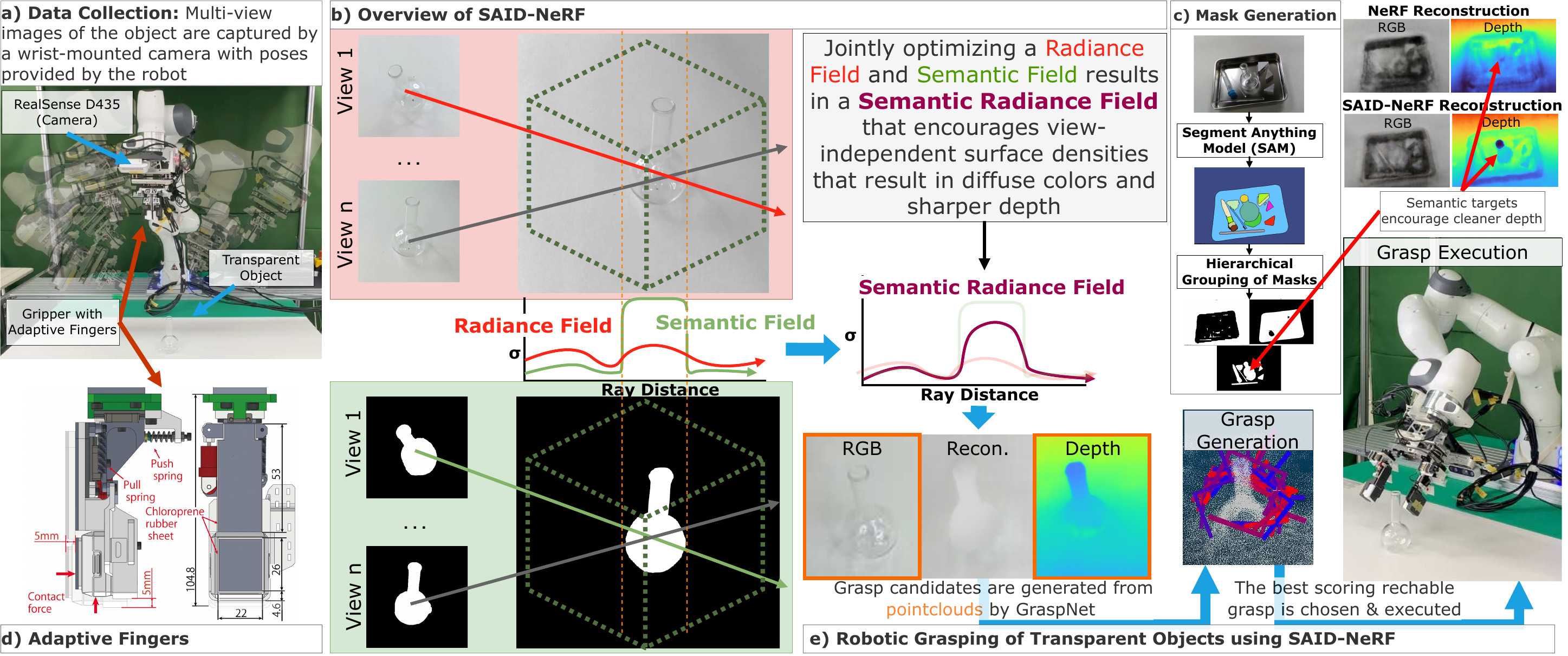}
    \caption{Overview of the proposed system for robotic grasping using \textbf{S}egmentation-\textbf{AID}ed \textbf{NeRF} (SAID-NeRF) for depth estimation of transparent objects. SAID-NeRF uses Visual Foundation Models (VFMs) for zero-shot segmentation to guide the reconstruction process}
    \label{fig:concept}
    \vspace{-6mm}
\end{figure*}

In this work, we propose a method \textbf{S}egmentation-\textbf{AID}ed \textbf{NeRF} (SAID-NeRF) that exploits instance masks generated by Visual Foundation Models (VFMs)~\cite{bommasani2021opportunities, Kirillov_2023_ICCV} to enhance NeRF's estimation of surfaces of transparent objects that are otherwise ambiguous when relying purely on RGB (Fig.~\ref{fig:concept}~(b)).
Having been trained on related visual tasks on internet-scale datasets for generalization, VFMs encounter a large variety of transparent objects under numerous conditions that can be leveraged.
In addition, for scenarios where objects in the scene are difficult to separate such as overlapping or cluttered objects, we present a simple heuristic to generate a hierarchy of non-overlapping masks for label-free use (Fig.~\ref{fig:concept}~(c)).
This allows the reconstruction of the objects in reasonable time (seconds) using a handful of views.
Then, the depth information acquired using the proposed method is used by the robot for object grasping.

The novelty and contributions of this work are 
\begin{itemize}
    \item Utilization of instance masks from VFMs to guide surface density acquisition of transparent objects.
    \item Extensions of NeRF that enable robust and reliable estimation under difficult conditions in seconds.
    \item A method for the grouping of label-less object segments into a hierarchical set of non-overlapping masks for the label-free generation. 
    \item Evaluation of the proposed method on a large-scale transparent object depth completion dataset.
    \item Evaluation of the proposed method through robotic grasping of several transparent objects under challenging settings.
\end{itemize}

\section{Related Work}
\label{sec:related_works}

\subsection{Transparent Object Segmentation}
\label{sec:relwork_perception}
Deep Learning models typically require large amounts of annotated data to learn, which is made challenging by the difficulty in acquiring ground truth masks and the diversity of appearance of transparent objects. The generation and use of synthetic data as an alternative is challenging due to the difficulty in accurate simulation of specular effects potentially inducing a sim2real gap.
Several datasets for transparent objects have been proposed for segmentation~\cite{xie2020segmenting}, depth completion and grasping~\cite{fang2022transcg,chen2022clearpose}. However, they remain limited in the diversity of views, settings, and objects. 

An alternative to methods trained on annotated data is the use of Visual Foundation Models (VFMs).
These models are tried on vast amounts of data leveraging self-supervision, other pre-trained models and combine modalities like text to aid scaling and generalization.
A VFM of particular interest to this work is the Segment Anything Model (SAM)~\cite{Kirillov_2023_ICCV}, notable for its 
 excellent zero-shot segmentation results. It is often combined with other methods to create powerful segmentation methods such as Grounded-SAM, Semantic-SAM~\cite{li2023semantic}, and SEEM~\cite{zou2023segment}.

In this study, we leverage SAM's zero-shot ability to segment transparent objects in wide contexts and conditions.

\subsection{Learning-based Depth Completion}
A class of prior works aims for the supervised completion of depth values trained entirely on synthetic data~\cite{sajjan2020clear, dai2022domain, ichnowski2021dex} or real data~\cite{chen2022clearpose, fang2022transcg} with depth ground truths often found by using 3d models or object markers~\cite{fang2022transcg}.
A common issue with supervised learning approaches is their generalization gap or the struggles of these models to work on data outside of their training set.
This is especially important for transparent objects since they demonstrate a wider range of environment-dependent appearance owing to their specular properties.
For a more comprehensive review of the topic, we refer the reader to~\cite{jiang2023robotic}. 
There also exists a wider range of monocular depth estimation methods that learn either relative or metric depth estimation~\cite{bhat2023zoedepth} and are trained on a large number of scenes for generalization but do not typically offer the millimeter precision or high performance on transparent objects required for successful grasping.

\subsection{Neural Radiance Fields (NeRFs)}
\label{sec:prior_nerf}
NeRFs~\cite{mildenhall2021nerf} have emerged as a promising 3d representation method to capture view-dependent appearance.
Several studies have looked at better representations such as Multiresolution Hash Encoding~\cite{mueller2022instant} and to improve reconstruction quality and time~\cite{martinbrualla2020nerfw, rebain2021derf, boss2021nerd, srinivasan2021nerv, zhang2021nerfactor, verbin2022ref}. Prior work has also explored using additional sources of supervision such as depth to reduce reconstruction time and views \cite{roessle2022dense} including using depth predicted from trained models \cite{wang2023sparsenerf, uy-scade-cvpr23, prinzler2023diner, song2023darf}.
Others have considered using NeRFs for semantic labeling~\cite{VoraRGMGSPTD22, kundu2022panoptic, zhi2021place, fu2022panoptic, mirzaei2023spin} and incorporating semantic information into NeRFs.
Methods that include semantic labels to objects in the scene typically assume we have a pre-trained NeRF, which is then used to convert 2d semantic masks into 3d~\cite{VoraRGMGSPTD22, kundu2022panoptic} or have a focus on label propagation~\cite{zhi2021place}.
This can be used for downstream tasks such as object selection, editing, in-painting etc.

By construction, \emph{radiance fields} aim to capture the appearance of the object without consideration of its true shape or depth. This problem is further exacerbated for transparent objects where surface specular effects may be perfectly captured without recovering the object's underlying geometry, rendering them unreliable for robotic tasks.
Implicit Surface estimation methods like NeUS \cite{wang2021neus, wang2023petneus} are better suited to this task but have longer training times, which is a bottleneck to tasks like robotic grasping.
Our proposed method seeks to use the NeRF formulation with semantic information (a Semantic-NeRF) for better surface extraction aimed at robotic grasping, ignoring effects on visual quality.

\subsection{Robotic Grasping with NeRF}
Several works have explored using NeRFs for grasping transparent objects.
Dex-NeRF~\cite{ichnowski2021dex} uses NeRFs to grasp transparent objects in the scene but required significant computation time in the order of hours.
The introduction of Multiresolution Hash Encoding~\cite{mueller2022instant} significantly shortened reconstruction time and has driven follow-up work like Evo-NeRF~\cite{kerr2022evo}.
GraspNeRF~\cite{dai2023graspnerf} uses a pre-trained feature extractor that is optimized jointly with a volumetric rendering loss (for RGB) and a trained volumetric grasp detection network VGN~\cite{breyer2021volumetric} to use grasp quality to guide detection of geometry and is trained end-to-end.
Our approach, instead, uses segmentation masks to overcome the view-dependent appearance of these objects enabling robotic grasping in a reasonable time using a handful of views. We provide a comparison to these methods and show the effectiveness of our proposed method in \Cref{sec:expt-results-robotic}.

\section{Method}
\label{sec:method}

This section describes the proposed method SAID-NeRF for depth completion of transparent objects as well as its proposed use in a robotic grasping system.
First, we describe the basics of NeRF and it's variants in the preliminaries (\Cref{sec:NeRF}) with a discussion of SAID-NeRF's purported working mechanism in the reconstruction of transparent objects (\Cref{sec:Semantic Fields}) as well as proposed extensions to the architecture (\Cref{sec:Position Encoding}) and the methodology to generate hierarchical layers of semantic masks of objects in the scene (\Cref{sec:maskgeneration}).
The use of SAID-NeRF in a robotic grasping system for the grasping of transparent objects is discussed in \Cref{sec:transparentgrasping}.

\subsection{Preliminary: Neural Radiance Fields (NeRFs)}
\label{sec:NeRF}
NeRFs~\cite{mildenhall2021nerf} use multilayer perceptions (MLPs) to represent a 3d volumetric scene with a 5d function of position $\textbf{x} = (x,y,z)$ and viewing direction $\textbf{d}= (\theta, \phi)$.
Specifically, for a given 3d point, NeRF learns an implicit function $\textbf{f}$ that estimates the density $\sigma$ and RGB color $\textbf{c}$ like $\textbf{f}(\textbf{x},\textbf{d})=(\sigma(\textbf{x}), \textbf{c}(\textbf{x},\textbf{d}))$.
To compute the color of a single pixel, we generate a ray $\mathbf{r}$ originating at the center of its projection direction.
The expected color $C(\mathbf{r})$ is calculated by integrating the implicit radiance field along this ray to compute the implicit radiance of any object within its path.
\begin{equation}
    C(\mathbf{r}) = \int_{t_{near}}^{t_{far}} T(t) \sigma(t) \mathbf{c}(t, d) dt,
    \label{eq:nerf_color}
\end{equation}
where $T(t) = \exp\left(-\int_{t_{near}}^{t}\sigma(s)ds\right)$ and $t_{near}$ and $t_{far}$ are the near and far planes of the camera.
NeRF approximates this integral using numerical quadrature along with techniques for efficient rendering~\cite{ mueller2022instant}. 

\begin{figure}[t]
    \centering
    \includegraphics[width=0.8\columnwidth]{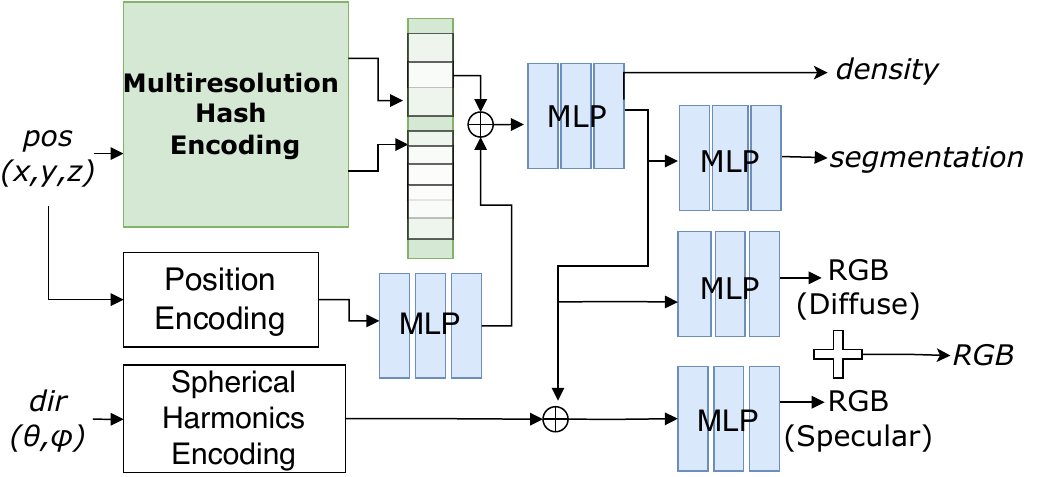}
    \caption{Architecture diagram of the proposed SAID-NeRF}
    \label{fig:architecture}
\end{figure}

Following~\cite{zhi2021place}, we augment the formulation by adding a segmentation rendering MLP before the viewing direction is appended for the RGB(Specular) component as shown in Fig.~\ref{fig:architecture}. The viewing direction itself is first projected using a Spherical Harmonics (SH) basis function before it is used.
Concretely, we introduce a view-invariant function $\textbf{s}$ that maps position $\textbf{x}$ to a distribution over $\textbf{C}$ semantic labels leading to 
\begin{equation}
    S(\mathbf{r}) = \int_{t_{near}}^{t_{far}} T(t) \sigma(t) \mathbf{s}(t) dt.
    \label{eq:nerf_semantic}
\end{equation}

Several extensions to NeRF aim to better capture specular effects by the decomposition of the color component $\textbf{c}$~\cite{verbin2022ref, zhang2021nerfactor}.
To better encourage the capture of the underlying diffuse color of transparent objects, we follow~\cite{tang2022nerf2mesh} in decomposing appearance into view-independent diffuse color $c_d(\textbf{x})$ and a view-dependent specular color $c_s(\textbf{x},\textbf{d})$ with shallow MLPs with the final color obtained as a sum $c = c_s+c_d$ as shown in Fig.~\ref{fig:architecture}, with a small L2 penalty on the specular term to encourage diffuse color components.

\subsection{Semantic Fields as NeRF Density Grounding}
\label{sec:Semantic Fields}
This section describes how joint optimization of semantic components in Semantic NeRFs can help recover better surface densities for transparent objects.
NeRF approximates \cref{eq:nerf_color}, and the expected color is given by
\begin{equation}
    \hat{C} = \sum_{i=1}^NT_i(1-\exp(-\sigma_i\delta_i))\mathbf{c}_i,
\end{equation}
where $T_i=\exp\left(-\sum_{j=1}^{i-1}\sigma_j\delta_j\right)$ and $\delta_i=t_{i+1}-t_i$ is the distance between two samples along the ray.
In a similar vein to the color estimation, the depth map in NeRF can be approximated using the transmittance \(T_i\) and volume density \(\sigma_i\) at each depth sample along the ray:
\begin{equation}
\hat{D} = \sum_{i=1}^N T_i (1 - \exp(-\sigma_i \delta_i)) t_i.
\label{eq:density_eq}
\end{equation}

When a projected ray encounters an opaque object at some distance $t_i$, it has a high contribution to the estimated color of the pixel color $\hat{C}$, acquired by a high assignment of volume density $\sigma_i$ at the point, therefore leading to high depth contribution $\hat{D}$ in \cref{eq:density_eq}.
However, when it encounters a transparent object, the contribution it has to the color of the pixel $\hat{C}$ is very low (visualized in the radiance field component of Fig.~\ref{fig:concept}~(b)).
Therefore, the assigned volume densities are much more diffuse until the ray passes through and perhaps encounters another opaque surface behind it.
This means that the values of estimated depth for the ray $\hat{D}$ will be closer to the opaque object behind it, or worse, somewhere in between, as is expected with an expectation of a bimodal distribution. 

In this work, we exploit this fact and leverage density assignments from the semantic head of the NeRF as a way to enforce higher surface densities to transparent objects in spite of the negative contribution they have to their color components.
Visualized in the Semantic Radiance Field component of Fig.~\ref{fig:concept}~(b), we see the visual effect of these density assignments in the diffuse color acquired by the transparent objects that were provided as masks, as a result of focus on view-independent, diffuse appearance.
This phenomenon is also what allows the reduction in the number of views and environment modifications like light sources required by existing methods, such as Dex-NeRF~\cite{ichnowski2021dex}.
\subsection{Position Encoding}
\label{sec:Position Encoding}
This section describes the addition of Frequency Embedding back into the architecture of a Multiresolution Hash Encoding-based NeRF for avoiding floaters and warped geometry for patternless surfaces.

Our proposed method is based on Instant-NGP~\cite{mueller2022instant}. Inspired by the work of ~\cite{zhu2023rhino}, we find the addition of the frequency embedding, by concatenation with the features of the multiresolution hash encoding, significantly improves the quality of reconstructions.
We hypothesize this to be an inductive bias towards preferring plane-like surface reconstructions due to positional encoding embedding regions with element-wise coordinates, loosely based on~\cite{wang2023petneus}.
This is particularly stark when objects are placed on unremarkable backgrounds that are non-patterned or single-colored leading to inaccurate densities and so-called ``floaters''.
These floaters have a particularly destructive effect on robotic grasping by confusing grasp generation and motion planning algorithms.

\subsection{Detection \& Semantic Mask Generation}
\label{sec:maskgeneration}
This section describes the detection of all objects in a scene using VFMs in a zero-shot manner and ways to generate roughly consistent semantic masks without the need for class labels.
To provide semantic targets to SAID-NeRF, an ordered set of $\textbf{S}$ semantic classes that are consistent across views are required.
Additionally, the segmentation method must be capable of separating transparent objects, both from their backgrounds and each other.
We propose a way to use SAM to generate unlabelled instance segments of an image and then use a straightforward heuristic to group them into a hierarchical set of non-overlapping scene masks that are roughly consistent as visualized in Fig.~\ref{fig:concept}~(c).

First, we make the following assumptions about the captured scene:
1) Some surface (e.g. a table) exists on which the object(s) of interest are placed.
2) Images capture this surface prominently (it occupies a significant part of the image) and consistently (present across multiple views).
3) Objects are placed on the surface directly or on top of/within each other, such as tools in a tray or water in a beaker. 

A simple algorithm is proposed to group masks into non-overlapping hierarchies, based on checking for overlap between their convex hulls as membership into non-overlapping sets (reminiscent of a union-find algorithm) as described in \Cref{alg:sam_to_set}.
We find that this algorithm, while straightforward, is quite imprecise and fails to handle several settings.
However, empirically, we find that it provides reasonably consistent masks as long as the number of sets requested remains reasonably low ($\leq 3$) and faulty assignments are typically compensated for by other views.

\begin{algorithm}[t]
\caption{Hierarchical semantic mask generation}
\label{alg:sam_to_set}
\algrenewcommand\algorithmicrequire{\textbf{Input:}}
\algrenewcommand\algorithmicensure{\textbf{Output:}}
    \begingroup
    \scalefont{0.85}
    \begin{algorithmic}[1]
        \Require $M$ masks, $c$ requested number of channels
        \Ensure $c$ masks of size $M$   
        \State Sort $M$ in decreasing order of area
        \State Compute convex hulls of each mask in $M$
        \State Initialize an empty list for the mask sets, $Sets$
        \For {each mask $i$ in sorted list of masks}
            \For {each set $s$ in $Sets$}
                \State check if hull of $i$ overlaps with hulls in set $s$
                \If {there is no conflict}
                    \State assign mask $i$ to set $s$
                \EndIf
            \EndFor
            \If {mask $i$ couldn't be added to existing set}
                \State create new set with $i$ and append it to $Sets$
            \EndIf
        \EndFor
        \If {$\mid Sets \mid > c$}
            \State merge all masks in $s \in Sets_{i>c}$ into $Sets_c$
        \EndIf
        
    \State \textbf{return} $Sets$
    \end{algorithmic}
    \endgroup
\end{algorithm}

\subsection{Robotic Grasp Planning}
\label{sec:transparentgrasping}
This section describes the use of SAID-NeRF in a robotic system for the grasping of transparent objects. A typical vision-based pipeline consists of perception (via RGB-D cameras), completion/estimation of depth, grasp generation, and grasp execution. We use a wrist-mounted RGB-D camera to automate data collection from different angles. By using camera poses acquired from the robot's pose, we overcome the need to use COLMAP~\cite{schonberger2016structure}, and operate in the robot's coordinate system making inter-operation easy and eliminating discarding images for lack of feature matches. 

Next, SAID-NeRF processes the collected images and pose information and reconstructs the scene in 3d. Depending on the format needed by the grasp generation (RGB-D, Point clouds, Mesh, Truncated Signed Distance Field, etc) the NeRF can flexibly generate the output. To generate point clouds, we accumulate rendered RGB and (estimated) depth values from the NeRF across training views. Point clouds are then used to generate potential grasp candidates using GraspNet~\cite{fang2020graspnet}, the highest-scoring reachable candidate of which is chosen and executed. While several grasp generation methods are applicable, we chose GraspNet~\cite{fang2020graspnet} motivated by its large-scale training, aim of generalization, as well as its use in several recent works~\cite{fang2022transcg, liu2022unseen,dai2022domain}.

\section{Experiment: Depth Completion Dataset}
\label{sec:expt_dataset}

\subsection{Experiment Setup}
\begin{figure}[t]
    \centering
    \includegraphics[width=1\columnwidth]{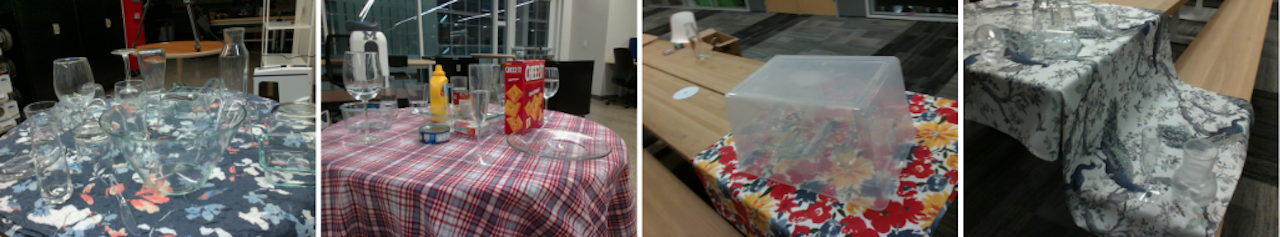}
    \caption{Samples from the ClearPose dataset~\cite{chen2022clearpose} in a cluttered setting, with opaque distractors, a translucent cover, and non-planar settings}
    \label{fig:datasetsamples}
\end{figure}
To evaluate our transparent object depth completion capabilities, we chose the ClearPose Transparent Object Dataset~\cite{chen2022clearpose} that consists of 63 objects arranged in numerous configurations.
The choice was motivated by the dataset's convenient categorization, its size, and the availability of pose information.
Specifically, we use the provided ``downscaled'' variant of their dataset, which subsamples images in each set and scene. Some sample images from the dataset are shown in Fig.~\ref{fig:datasetsamples}.
The various test sets explore a variety of settings, and noticeably, some have as few as 14-23 views taken from a wide range of viewing angles.
We show that this does not hinder the proposed method and can perform competitively even in these difficult sparse view conditions.

As baseline methods, we report the results of the depth completion method, TransCG~\cite{fang2022transcg}, presented in the original work as a baseline.
Additionally, we also benchmark the recently proposed depth completion method DREDS~\cite{dai2022domain} trained on a mix of transparent, metallic, rough, and specular objects. We note that we do not perform any fine-tuning of the methods on the training sets.
To evaluate against multi-view NeRF-based approaches, we also benchmark against the \emph{Nerfacto}~\cite{tancik2023nerfstudio} method which combines several recent advances in NeRF literature.
Additionally, to evaluate the use of large-scale monocular depth prediction methods in combination with NeRF, we benchmark against a variant of Nerfacto called the \emph{Depth-Nerfacto} which uses the monocular depth estimation method ZoeDepth~\cite{bhat2023zoedepth} and a Ranking-based depth loss from SparseNerf~\cite{wang2023sparsenerf} to compensate for the unknown scale and metric of depth values generated which prevent it from being usable directly.

\subsection{Implementation Details}
We implement SAID-NeRF using the Nerfacc~\cite{Li_2023_ICCV} library for efficient sampling and use their proposal network estimator, first proposed in the Mip-NeRF 360 paper~\cite{barron2022mipnerf360}.
The base architecture is implemented as per Fig.~\ref{fig:architecture} with the additional position encoding MLP of a single layer of width 16, while the segmentation MLP is two layers of width 32.
The loss function for the RGB values (output as a sum of the diffuse and specular heads) is chosen to be the L1 loss.
For segmentation targets, we generate three sets of masks (such that the segmentation head outputs a vector of length 3). For their quick generation, we use a ``light'' variant of HQ-SAM (Called Light-HQ SAM)~\cite{sam_hq}.
The semantic labels are jointly trained along with RGB using a Binary cross-entropy loss.
We find that depth supervision imposed on the expected depth values was vital for scenes where backgrounds did not possess significant patterns. We use an L2 loss imposed on the depth values obtained from the RGB-D camera (reminiscent of \cite{roessle2022dense}) despite its inability to capture depth values for transparent objects.

We then allow the optimization to run on the dataset and cap the training steps to a specific time limit $T$, which is 20 seconds for the robotic experiments and 60 seconds for ClearPose.
We note that this runs counter to the step-based counting that is usually reported but is more meaningful in an application environment, as the hyperparameters and scene can significantly impact the computation time of a single step.
We run all the methods described here on an NVIDIA RTX4000 Ada GPU, except the SAID-NeRF results for the ClearPose Transparent Object Depth Completion dataset, which were run using an NVIDIA Tesla V100 GPU.

\subsection{Ablation Study \& Data Requirements}
In this section, we ablate and evaluate the performance of each component of the proposed method described in \Cref{sec:method}.
We systematically remove semantic supervision, position encoding, and depth supervision to study their contribution. Additionally, we report a vanilla variant of Instant-NGP as a baseline without any of the proposed extensions, trained for $10\times$ longer. For evaluation, we choose the \emph{Heavy Occlusion} subset of the ClearPose dataset~\cite{chen2022clearpose}.

Table~\ref{tab:results_clearpose_ablation} shows the results.
We find that the components with the largest impact are depth supervision followed by position encoding.
Both of these have significant regularizing effects on the geometry of the scene as a whole and make the difference between a coherent scene geometry.
Semantic information and the Diffuse-Specular split, as expected, have a small effect since they are both extensions proposed to aid the capture and reconstruction of transparent objects and have little to do with opaque objects, which make up a big portion of the scene.
Additionally, we find that the proposed method SAID-NeRF, with all its proposed extensions, can significantly outperform vanilla Instant-NGP using as few as four images (and evaluating on all the available views) or training for under 10 seconds (Table~\ref{tab:results_clearpose_dsetsize} and Table~\ref{tab:results_clearpose_traintime}).

\begin{table}[t]
    \centering
    \begingroup
    \scalefont{1.0}
    \caption{Ablation study for the various components proposed and discussed in \Cref{sec:method}}
    \begin{tabular}{lrrr}
        \toprule
        Method & RMSE & MAE & REL \\
        \midrule
        SAID-NeRF (Full) (ours)  & $\bf{0.0843}$ & $\bf{0.0599}$ & $\bf{0.0698}$ \\
        SAID-NeRF  w/o Position Encoding & 0.1890 & 0.1013 & 0.1202 \\
        SAID-NeRF  w/o Semantic Info. & 0.1308 & 0.0832 & 0.0964 \\
        SAID-NeRF  w/o Diffuse/Specular & 0.0921 & 0.0675 & 0.0788 \\
        SAID-NeRF  w/o Depth Supervision & 0.6328 & 0.6100 & 0.7854 \\
        Instant-NGP & 0.2051 & 0.1754 & 0.2250 \\
        \bottomrule
    \end{tabular}
    \label{tab:results_clearpose_ablation}
    \endgroup
    \vspace{2mm}
\end{table}

\begin{table}[t]
    \centering
    \begingroup
    \scalefont{0.9}
    \setlength\columnsep{4pt}
    \begin{multicols}{2}
    \caption{SAID-NeRF results on reducing the number of views for the \textbf{Heavy Occlusion} dataset} %
    \begin{tabular}{@{}c|rrr}
        \toprule
        Views & RMSE & MAE & REL\\
        \midrule
        All & 0.0837 & 0.0586 & 0.0673\\
        60 & 0.0828 &  0.0587 & 0.0669\\
        30 &  0.0912 & 0.0655 & 0.0733\\
        15 &  0.1092 & 0.0794 & 0.0873\\
        8 & 0.1196 & 0.0883 & 0.0971  \\
        4 &  0.1298 & 0.0981 & 0.1088 \\
        \bottomrule
    \end{tabular}
    \label{tab:results_clearpose_dsetsize}
    \caption{SAID-NeRF results on reducing the NeRF reconstruction time for the \textbf{Heavy Occlusion} dataset (all views)}

    \begin{tabular}{@{}c|rrr}
        \toprule
        Train&&&\\
        Time & RMSE & MAE & REL \\
        \midrule
        10s &  0.1025 & 0.0753 & 0.0884 \\
        15s &  0.0993 & 0.0737 & 0.0846	\\
        30s &  0.0877 & 0.0626 & 0.0721 \\
        60s &  0.0828 & 0.0587 & 0.0669 \\
        90s &  0.0842 & 0.0610 & 0.0702 \\
        \bottomrule
    \end{tabular}
    \label{tab:results_clearpose_traintime}
    \end{multicols}
    \endgroup
    \vspace{-4mm}
\end{table}

\begin{figure*}[t]
    \centering
\includegraphics[width=2\columnwidth]{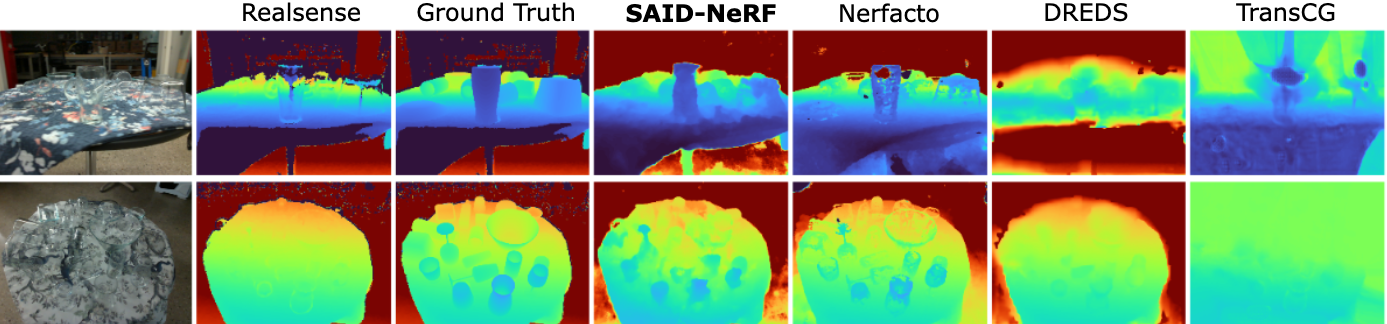}
    \caption{Visualization of completed depth on scenes from ``New Background'' and ``Heavy Occlusion'' from the ClearPose~\cite{chen2022clearpose} dataset}
    \label{fig:clearpose_results}
    \vspace{-4mm} 
\end{figure*}
\subsection{Experiment Results}
The results of the methods, including the proposed SAID-NeRF (ours), are presented in Table~\ref{table:results_clearpose}.
The evaluation protocol and metrics closely follow that set by~\cite{sajjan2020clear}:
The Root Mean Squared Error in meters (RMSE), the median error relative to the depth (REL), and percentages of pixels with predicted depths falling within an interval $([\delta=|predicted-true|/true]$ where $\delta \in \{1.05, 1.10, 1.25 \}$.
The error is averaged first across the scene and then within the dataset for the fairness of multi-view methods.

We see that the proposed SAID-NeRF outperforms TransCG, DREDS, and Nerfacto in most of the test sets. 
Both the ``New background'' and ``Heavy occlusion'' test sets have between 67-89 images, but the performance differences between them across all methods are remarkable.
The ``Filled liquid'', ``Opaque distractor'', ``Translucent cover'' and ``Non-planar'' datasets only have between 14-23 images.
Remarkably, we see that the performance of SAID-NeRF remains relatively consistent despite this drop in size by a factor of 3 and increased scene complexity. Although DREDS outperformed SAID-NeRF on the ``Translucent cover'' set, the difference is not significant and its remarkable performance across all sets is roughly consistent.
This stability in performance may be highly sought after in applications that require a large degree of generalization.
We also note that DREDS, which was trained not just for transparent objects but also for rough, specular, and shiny objects, is more stable throughout.
A visualization of completed/estimated depth values from the methods on 3 scenes from the ``New Background'' and ``Heavy Occlusion'' datasets are presented in Fig.~\ref{fig:clearpose_results}.
Although DREDS benchmarked well (in terms of RMSE), from the visualization we find it did not accurately estimate object depths. The proposed method is able to accurately estimate the depth of objects, even transparent objects, without significant loss of shape.

\begin{table}[t]
    \centering
    \begingroup
    \scalefont{1.0}
    \caption{Results on the test set of the ClearPose dataset. RMSE, REL, and MAE are reported in meters}
    \begin{tabular}{@{}cc|@{}c|cccc@{}c@{}c@{}}
\toprule
&& Method & RMSE & REL & MAE & \,$\delta_{1.05}$ & \,$\delta_{1.10}$ & \,$\delta_{1.25}$ \\
\midrule

\parbox[c]{2mm}{\multirow{5}{*}{\rotatebox[origin=c]{90}{\begin{tabular}{c}\scriptsize{New}\\ \scriptsize{background}\end{tabular}}}}
&& \textbf{SAID-NeRF(Ours)} & $\bf{0.0724}$ & $\bf{0.0609}$ & $\bf{0.0530}$ & 54 & $\bf{81}$ & $\bf{97}$ \\
&& Nerfacto & 1.3400 & 0.2200 & 0.2300 & $\bf{63}$ & 79 & 91 \\
&& Nerfacto (ZoeDepth) & 0.6253 & 0.1860 & 0.1705 & $\bf{63}$ & 79 & 91 \\
&& TransCG & 0.1577 & 0.1206 & 0.1177 & 31 & 53 & 81 \\
&& DREDS & 0.1406 & 0.1236 & 0.1032 & 34 & 59 & 87 \\
\hline

\parbox[c]{2mm}{\multirow{4}{*}{\rotatebox[origin=c]{90}{\begin{tabular}{c} \scriptsize{Heavy}\\\scriptsize{occlusion}\end{tabular}}}}
&& \textbf{SAID-NeRF(Ours)} & $\bf{0.0843}$ & $\bf{0.0698}$ & $\bf{0.0599}$ & 48 & $\bf{76}$ & $\bf{96}$ \\
&& Nerfacto & 0.5800 & 0.2100 & 0.2200 & 50 & 62 & 72 \\
&& Nerfacto(ZoeDepth)& 0.7160 & 0.2073 & 0.1952 & $\bf{54}$ & 71 & 84 \\
&& TransCG & 0.1769 & 0.1371 & 0.1395 & 30 & 53 & 82 \\
&& DREDS & 0.1770 & 0.1624 & 0.1281 & 26 & 47 & 81 \\
\hline

\parbox[c]{2mm}{\multirow{4}{*}{\rotatebox[origin=c]{90}{\begin{tabular}{c} \scriptsize{Filled}\\\scriptsize{liquid}\end{tabular}}}}
&& \textbf{SAID-NeRF(Ours)} & $\bf{0.0754}$ & $\bf{0.0709}$ & $\bf{0.0581}$ & 46 & $\bf{74}$ & $\bf{96}$ \\
&& Nerfacto & 0.3300 & 0.2300 & 0.2700 & 43 & 52 & 58 \\
&& Nerfacto(ZoeDepth)& 0.4154  & 0.2797 & 0.2356 & $\bf{47}$ &  56 & 62 \\
&& TransCG & 0.1096 & 0.1024 & 0.0867 & 34 & 59 & 88 \\
&& DREDS & 0.1486 & 0.1387 & 0.1112 & 24 & 46 & 87 \\
\hline

\parbox[c]{2mm}{\multirow{4}{*}{\rotatebox[origin=c]{90}{\begin{tabular}{c} \scriptsize{Opaque}\\ \scriptsize{distractor}\end{tabular}}}}
&& \textbf{SAID-NeRF(Ours)} & $\bf{0.1020}$ & $\bf{0.0890}$ & $\bf{0.0749}$ & $\bf{41}$ & $\bf{65}$ & $\bf{91}$ \\
&& Nerfacto & 0.5800 & 0.2800 & 0.2900 & 37 & 43 & 48 \\
&& Nerfacto(ZoeDepth)& 0.6510 & 0.3214 & 0.2946  & 40 & 46 & 52 \\
&& TransCG & 0.1815 & 0.1456 & 0.1347 & 29 & 47 & 75 \\
&& DREDS & 0.1418 & 0.1211 & 0.0946 & 40 & 58 & 87 \\
\hline

\parbox[c]{2mm}{\multirow{4}{*}{\rotatebox[origin=c]{90}{\begin{tabular}{c}\scriptsize{Translucent}\\\scriptsize{cover}\end{tabular}}}}
&& \textbf{SAID-NeRF(Ours)} & 0.1173 & 0.1152 & 0.0989 & 25 & 46 & 84 \\ 
&& Nerfacto & 0.3800 & 0.3300 & 0.3700 & 19 & 26 & 37 \\
&& Nerfacto(ZoeDepth)& 0.3222 & 0.3120 & 0.2650 & 21 & 29 & 45 \\
&& TransCG & 0.1368 & 0.1278 & 0.1132 & 27 & 48 & 78 \\
&& DREDS & $\bf{0.1102}$ & $\bf{0.1007}$ & $\bf{0.0831}$ & $\bf{33}$ & $\bf{59}$ & $\bf{90}$ \\
\hline

\parbox[c]{2mm}{\multirow{4}{*}{\rotatebox[origin=c]{90}{\begin{tabular}{c} \scriptsize{Non}\\ \scriptsize{planar} \end{tabular}}}}
&& \textbf{SAID-NeRF (Ours)} & $\bf{0.1116}$ & $\bf{0.0936}$ & $\bf{0.0843}$ & 38 & $\bf{64}$ & $\bf{93}$ \\
&& Nerfacto & 0.3300 & 0.2600 & 0.2600 & 40 & 52 & 63 \\
&& Nerfacto(ZoeDepth)& 0.3269 & 0.2164  & 0.2247  & $\bf{45}$ & 59  & 71 \\
&& TransCG & 0.2193 & 0.1649 & 0.1805 & 27 & 45 & 74 \\
&& DREDS & 0.1559 & 0.1347 & 0.1182 & 31 & 51 & 85 \\
\bottomrule
\end{tabular}

    \label{table:results_clearpose}
    \endgroup
\end{table}

\section{Experiment: Robotic Grasping}
\label{sec:expt_robotic}
To evaluate the applicability of SAID-NeRF as part of a robotic grasping system of transparent objects, we perform robotic experiments. First, we describe the setup of the experiment as well as present and discuss it's results.

\subsection{Experiment Setup}
\label{sec:robot_setup}

Our robotic system, shown in Fig.~\ref{fig:concept}~(e), consists of a Franka Emika Panda Arm 7-DoF robotic arm equipped with a custom-made parallel gripper operated by a servomotor (Dynamixel XM430-W350-R). A wrist-mounted Intel RealSense D435 is used to capture RGB-D images for data collection.
The parallel gripper is equipped with adaptive fingers for $5~\mathrm{\,mm}$ adjustment in the horizontal and vertical directions to adapt to errors and for gentle post-grasp placing.
The insides of the fingers are lined with a chloroprene rubber sheet to increase the coefficient of friction.

\begin{figure*}[t]
    \centering
    \includegraphics[width=1.8\columnwidth]{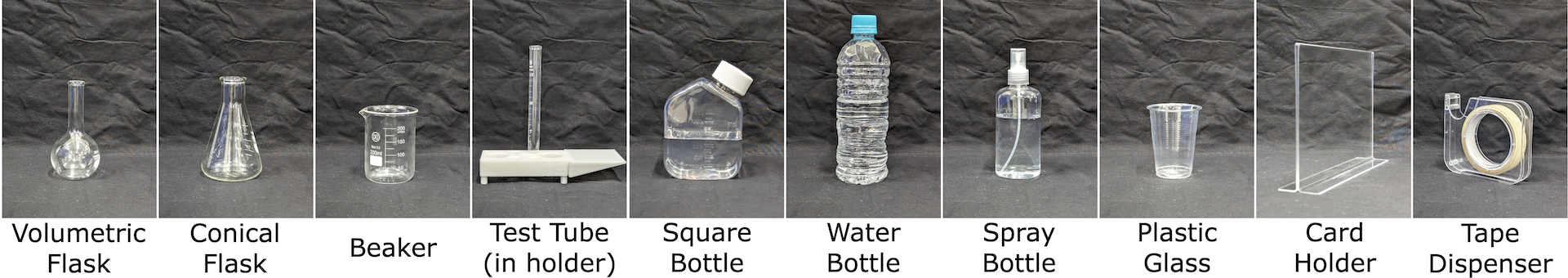}
    \caption{Transparent objects used for robotic grasping experiments}
    \label{fig:grasp_objects}
    \vspace{-1mm}
\end{figure*}

For the grasp targets, we select 10 objects shown in Fig.~\ref{fig:grasp_objects}.
The objects were chosen amongst laboratory and household equipment to ensure they span a large range of shapes, sizes, and surface gloss such that the difficulties in acquiring depth information would vary.
Objects are placed on a white table that is capable of reflecting light (Fig.~\ref{fig:concept}~(a)).
Additionally, a single strong directional light was present, which further confused methods that rely on vision. 

\subsubsection{Data Collection}
The first step of the process is the collection of datasets.
Each object is placed at the center of the workspace, and the robot samples 90 equidistant poses in the hemispheres of radii $[0.4-0.6]~\mathrm{\,m}$ around the object, visualized in Fig.~\ref{fig:concept}~(a).
Using the MoveIt!~\cite{coleman2014reducing} motion planner, the robotic arm reaches the target pose and records the RGB, depth, and corresponding pose.

\begin{table*}[]
\centering
\caption{Robotic grasping results for transparent objects}
\begin{tabular}{l|cccccccccc}
\hline
Method    & Volumetric  & Conical     & Beaker      & Test Tube   & Square      & Water        & Spray        & Plastic      & Card        & Tape        \\
          & Flask       & Flask       &             & (in holder) & Bottle      & Bottle       & Bottle       & Glass        & Holder      & Dispenser   \\ \hline
\textbf{SAID-NeRF (Ours)}      & $\bf{9/10}$ & $\bf{9/10}$ & $\bf{9/10}$ & $\bf{8/10}$ & $\bf{7/10}$ & $\bf{10/10}$ & $\bf{10/10}$ & $\bf{10/10}$ & $\bf{9/10}$ & $\bf{9/10}$ \\
Nerfacto  & $0/10$      & $3/10$      & $0/10$      & $7/10$      & $1/10$      & $2/10$       & $7/10$       & $0/10$       & $0/10$      & $0/10$      \\
GraspNeRF & $7/10$      & $8/10$      & $5/10$      & $1/10$      & $6/10$      & $4/10$       & $\bf{10/10}$ & $1/10$       & $1/10$      & $6/10$      \\ 
TransCG   & $5/10$      & $3/10$      & $4/10$      & $2/10$      & $2/10$      & $4/10$       & $4/10$       & $0/10$       & $4/10$      & $2/10$      \\
DREDS     & $5/10$      & $2/10$      & $3/10$      & $5/10$      & $1/10$      & $2/10$       & $5/10$       & $0/10$       & $5/10$      & $2/10$      \\ \hline
\end{tabular}
\label{table:res_robot}
\vspace{-6mm}
\end{table*}

\subsubsection{Grasping Process}

After data collection, point clouds are generated using depths completed/estimated by baselines and the proposed method. Using GraspNet~\cite{fang2020graspnet}, we generate and choose the highest scoring, reachable grasp to execute.
For each object, we execute 10 grasp attempts and count a success whenever an object is successfully lifted above the table height and is maintained for $\approx5$ seconds. This is done to verify a robust grasp since the objects being grasped are highly slippery and often irregularly shaped.

\subsubsection{Comparison Methods}
For our comparative analysis, we use two-depth completion methods from previous sections, DREDS~\cite{dai2022domain} and TransCG~\cite{fang2022transcg}, without any additional fine-tuning, and the NeRF method Nerfacto~\cite{tancik2023nerfstudio}, in combination with GraspNet for grasp generation. We also compare against GraspNeRF~\cite{dai2023graspnerf}, a high-performance transparent and specular object grasping method with its own grasp generation algorithm VGN~\cite{breyer2020volumetric}.

For our proposed SAID-NeRF and Nerfacto, we first create 10 sets of 30 images uniformly sampled from a collected 90 which is used to create the NeRF and generate 10 point clouds.
This is done to train and compare the methods under similar environmental conditions.
For DREDS and TransCG, we generate point clouds by combining 3-5 sampled views with their completed depth maps. We note that accumulating more views led to significant error accumulation. All point clouds are then cleaned via statistical outlier removal.
GraspNeRF uses 30 input views to generate a Truncated Signed Distance Field (TSDF), and subsequently, grasp candidates.

\subsection{Experiment Results}
\label{sec:expt-results-robotic}
\begin{figure}[t]
    \centering
\includegraphics[width=1\columnwidth]{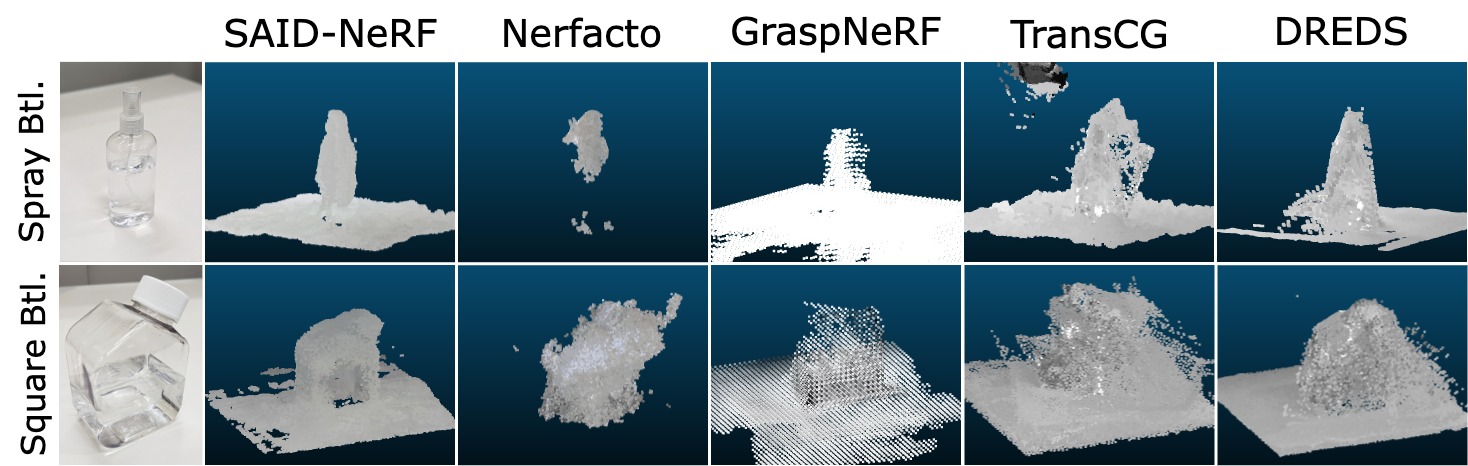}
    \caption{Point clouds generated by methods tested. GraspNeRF does not use point clouds and its TSDF was converted for visualization}
    \label{fig:pointcloud_results}
\end{figure}

The success rates (out of 10) of the various methods of grasping the transparent objects are presented in Table~\ref{table:res_robot}.
We see that point clouds generated by our proposed method lead to significant grasp performance across all the objects tested with some representative samples visualized in Fig.~\ref{fig:pointcloud_results}. 

For TransCG, we find that a majority of failure cases involve attempting to grasp noisy pixels that float around in the air despite having run outlier removal. More aggressive processing led to the loss of already challenging point density on the objects.
The performance of DREDS, while resulting in similar results to TransCG, is characteristically different.
It makes more reasonable grasp attempts but fails when the objects are accidentally pushed away due to the propensity of DREDS to predict smooth depth values along object boundaries in our scenes, especially against the shiny white table, leading to spurious points on its non-camera-facing side.
This is exacerbated by our views being sampled above the object, leading to the accumulation of spurious conical structures around the base of the objects, eliminating a significant number of stable side-ward grasps. 

Nerfacto, while able to render reasonable RGB shows highly variable behavior in terms of density. It shows reasonable performance with Spray Bottle and Test-tube (with holder), likely due to their opaque cap and base respectively helping with estimated density, while it fails on all other objects. 
GraspNeRF shows better performance amongst the methods tested and generates grasps different from those from GraspNet.
It excels at repeatable grasps for smaller objects such as Spray Bottles and Flasks, while its failure cases appear to be cascading and consistent for objects with low height (Plastic Glass and Beaker) or large volumes (over $30 ~\mathrm{\,cm^2}$) such as the Water Bottle, Card Holder and Test Tube (in holder).
We posit this is due to the voxel-based representation used by GraspNeRF not being able to sufficiently capture these objects within its bounds. Using larger values is taxing on GPU memory and causes a loss of details for smaller objects leading to worse grasps overall.

Our proposed method, SAID-NeRF, has the highest success rate on all $10/10$ objects and finds that the main failure modes are for the Square Bottle which is the heaviest object tested, and causes it to slip due to being picked up by it's cap, or by being accidentally pushed away during approach.

We note that for segmentation targets, two sets of masks are generated since the scenes are relatively simple.

\section{Discussion}
Our proposed method, SAID-NeRF, is able to do depth completion of a wide range of challenging transparent objects.
Limitations of our approach include the assumptions made by the hierarchical mask generation. While reasonable for typical indoor scenes, they may not transfer to other complex settings.
A workaround is using models that provide semantic labels, as briefly mentioned in \Cref{sec:maskgeneration}.
Another potential issue is the propensity of the semantic components to induce convexity to reconstructed shapes, causing objects with hollow surfaces or fine surface details to be lost. However, objects that can be grasped from the outer edges remain unaffected.
While beyond the aim of this work, these details may be recovered by a second stage of fine-tuning without the semantic component.

\section{Conclusion}
\label{sec:conclusion}
We introduce Segmentation-Aided NeRF, or SAID-NeRF, an extension of Semantic-NeRFs that uses semantic masks to guide depth estimation for transparent objects.
We use Instance Segmentation VFMs in a zero-shot way combined with a simple heuristic to automatically construct semantic masks that the NeRFs use to better recover object surfaces.
We find that our method is able to perform well under difficult environmental conditions due to its proposed extensions and outperforms several recent depth completion models trained on large datasets for transparent objects.
SAID-NeRF is also able to consistently outperform a strong NeRF method (Nerfacto) on all these tasks.
We further evaluate them for robotic grasping of transparent objects under challenging settings and additionally outperform GraspNeRF, a state-of-the-art NeRF-based grasping method that consists of trained and learnable components.

\section*{ACKNOWLEDGMENT}\small
The authors thank Dr. Sosuke Kobayashi, Dr. Shin-ichi Maeda, and Prof. Tadahiro Taniguchi for the many discussions.

\bibliographystyle{IEEEtran} %
\bibliography{IEEEabrv, main}

\end{document}